%% file: iclr2027_conference.tex
\documentclass{article}

\usepackage{float}
\usepackage{iclr2027_conference}
\usepackage{times}

\input{math_commands.tex}

\usepackage{amsmath}
\usepackage{amssymb}
\usepackage{graphicx}
\usepackage{xcolor}
\usepackage{booktabs}
\usepackage{array}
\usepackage{ragged2e}
\usepackage{algorithm}
\usepackage{algpseudocode}
\usepackage{hyperref}
\usepackage{url}

\newcolumntype{L}[1]{>{\RaggedRight\arraybackslash}p{#1}}

\definecolor{linkpink}{RGB}{220, 20, 120}

\newcommand{\projectlink}[3]{%
    \raisebox{-0.15\height}{%
        \includegraphics[height=1.15em]{#1}%
    }%
    \hspace{0.45em}%
    \href{#2}{\textcolor{linkpink}{\texttt{#3}}}%
}

\title{
\makebox[\linewidth][c]{%
    \raisebox{-0.22\height}{%
        \includegraphics[height=1.2em]{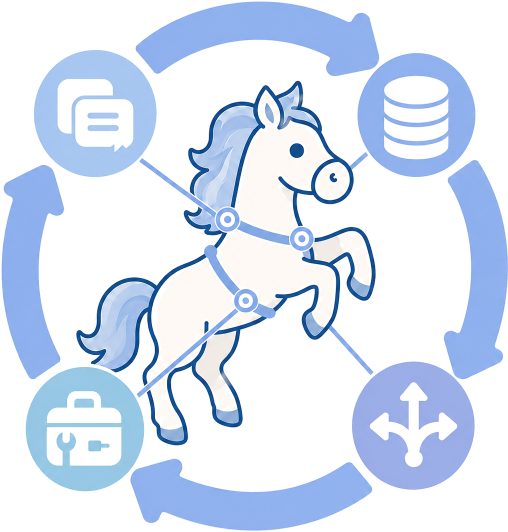}%
    }%
    \hspace{0.3em}%
    Harness Continual Learning: Continual%
}\\[0.10em]
\makebox[\textwidth][c]{%
    \hspace{-2.50em}%
    Adaptation Beyond Model Parameters%
}
}

\author{
Borui Kang$^{1}$,
Jinrui Gu$^{1}$,
Junhan Lv$^{1}$,
Wenbin Li$^{1}$\thanks{Corresponding author},
Lei Wang$^{2}$,
Yang Gao$^{1}$
\\[0.5em]
$^{1}$State Key Laboratory for Novel Software Technology,
Nanjing University, China
\\
$^{2}$University of Wollongong, Australia
}

\iclrdisplaycopy

\begin{document}

\maketitle


%
%
%
%
%

\begin{center}
\small

\projectlink
    {Figures/logo.pdf}
    {https://boringkey.github.io/Harness-Continual-Learning/}
    {Project Page}
\qquad
\projectlink
    {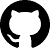}
    {https://github.com/RL-MIND/Harness-Continual-Learning}
    {Github}
\qquad
\projectlink
    {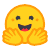}
    {https://huggingface.co/RL-MIND/Harness-Continual-Learning}
    {Hugging Face}

\end{center}

\vspace{0.6em}

\begin{abstract}
Continual learning has largely been model-centric, treating model parameters as
the state that changes with sequential experience. Modern agents can also adapt
through a harness of prompts, memories, tools, skills, and routing rules. Because
these contents jointly shape later execution, a harness update can disrupt previously
reliable behavior even when the model is frozen. This raises a new question:
how can an agent continually improve its state outside the model while retaining
behavior acquired earlier? We formulate \emph{Harness Continual Learning} (HCL), a
new continual learning paradigm in which the harness evolves around a frozen
foundation model, and define the resulting loss of earlier behavior as \emph{harness-level
forgetting}. We instantiate HCL with four execution-facing components: the
Task Interface, Experience Memory, Capability Map, and Adaptive Router. We
further introduce \emph{guarded harness evolution} to separate update generation from
state commitment. A Continual Optimizer proposes candidate harnesses from
post-execution feedback, and a Continual Evaluator commits the resulting candidate
harness only after checking current improvement, historical retention, and
validity. Experiments on open-world capability accumulation, textual reasoning, and
multimodal perception demonstrate capability accumulation and failure recovery,
with relative gains exceeding 10\% over corresponding baselines in multiple settings. Additional
experiments further examine knowledge transfer through harness evolution. Component ablations assess the contribution of
each harness component, while controlled retention sweeps reveal measurable
harness-level forgetting and demonstrate a controllable stability--plasticity trade-off. Our project is available at: \href{https://boringkey.github.io/Harness-Continual-Learning}{\textcolor{linkpink}{https://boringkey.github.io/Harness-Continual-Learning}}.
\end{abstract}

\section{Introduction}
\label{sec:introduction}

Continual learning studies how a system acquires capabilities from sequential experience while retaining previously learned behavior \citep{delange2021continual,shi2024continuallearninglargelanguage,lu2026little}. Existing formulations realize this process mainly by changing model parameters, representations, or architectural components; we refer to this established view as \emph{model-centric continual learning}.

\begin{figure*}[t]
\centering
\includegraphics[width=\linewidth]{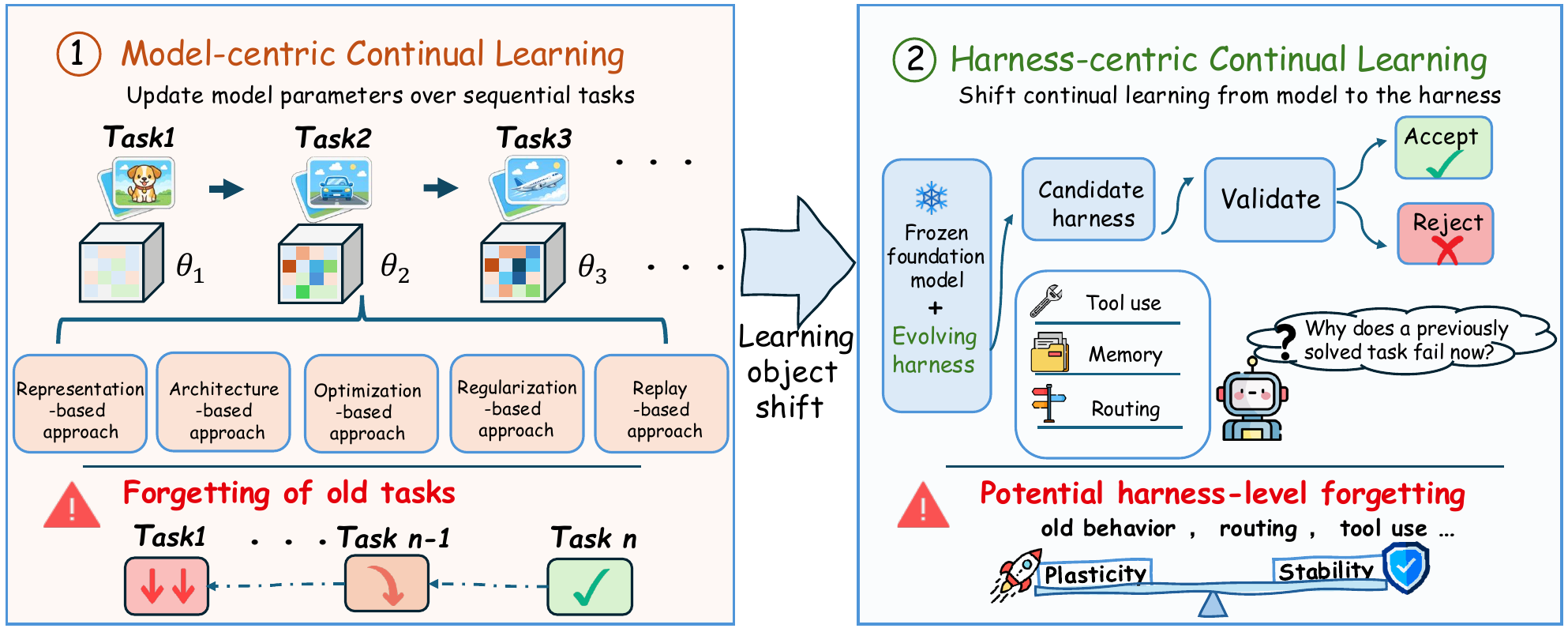}
\caption{The shift in the object of continual learning. Model-centric methods
update model parameters $\theta$ over sequential experience. HCL instead
updates harness state around a frozen foundation model. In both settings, adaptation can improve performance on new capabilities, but may also degrade capabilities acquired earlier.}
\label{fig:hcl_concept}
\end{figure*}

The rise of agentic AI introduces another source of adaptation: an external \emph{harness} that determines how a foundation model receives information, retrieves experience, and acts \citep{chen2026standalonellmsintegratedintelligence,
jimenez2024swebenchlanguagemodelsresolve,
li2026agentharness,meng_agent_2026_harness,
xie2024osworldbenchmarkingmultimodalagents,
xu2025theagentcompanybenchmarkingllmagents}. Prompts, memories, tool and skill specifications, and routing policies can persist and evolve across interactions even when the foundation model remains frozen. Agent adaptation is therefore no longer confined to model state: harness state can accumulate experience and reshape future behavior. This makes the harness a new object of continual learning research, as illustrated in Figure~\ref{fig:hcl_concept}.

We formalize this setting as \emph{Harness Continual Learning} (HCL), a continual learning paradigm that acquires and retains capabilities by sequentially updating harness state around a frozen foundation model.
Conventional harness optimization typically searches for prompts, functions, or workflows that improve a current objective
\citep{zhang2024agentoptimizer,zhang2025aflowautomatingagenticworkflow}; HCL instead studies a sequence of deployed harnesses. Because harness components are coupled in execution, a memory update can change evidence retrieved for an
earlier query, a skill revision can alter tool use, and a routing edit can break a previously successful workflow. An update that helps recent cases can therefore turn an earlier correct answer, valid tool call, or successful action trajectory
into a failure without changing the foundation model. We call this phenomenon \emph{harness-level forgetting}, extending the classical stability--plasticity problem from model state to harness state.

To study continual adaptation under this retention requirement, we develop an HCL framework with two parts. First, we define the Task Interface, Experience Memory, Capability Map, and Adaptive Router as the harness state and learning object of HCL. 
These components are jointly versioned and determine how the agent processes information, reuses experience and capabilities, and organizes execution. 
Second, guarded harness evolution governs state transitions through two modules: a Continual Optimizer that proposes candidate harnesses from post-execution feedback, and a Continual Evaluator that determines whether those candidates can be committed. 
Only a candidate harness that improves current validation performance
while satisfying the historical-retention budget and validity constraints can be committed as the deployed state. 
This proposal–evaluation–commitment process makes retention an explicit condition of harness adaptation, mitigating harness-level forgetting while controlling the stability–plasticity trade-off.

We evaluate HCL across textual reasoning, multimodal perception, and open-world interaction, and further study retention control, harness-level distillation,  and component contributions. In this work, our contributions are as follows:
\begin{itemize}
\item We propose and formalize \emph{Harness Continual Learning}, as a new continual learning paradigm, shifting the continual learning object from model state to harness state around a frozen foundation model.
\item We identify \emph{harness-level forgetting} and introduce \emph{guarded harness evolution}, which makes current improvement, historical retention, and validity explicit conditions for committing a harness update.
\item We show across reasoning, multimodal, and interactive streams that harness evolution supports capability accumulation. Ablations and retention sweeps further reveal component contributions, measurable forgetting, and controllable stability--plasticity trade-offs, with exploratory evidence for knowledge transfer.
\end{itemize}

\section{Related Work}
\label{sec:related_work}

\subsection{Harness Engineering}
\label{sec:harness_structure}

Contemporary agent systems place a runtime harness around a foundation model to turn inference into task-directed execution \citep{he_harness_2026,li2026agentharness,
meng_agent_2026_harness,zhou2026harness}. Harness state commonly includes input interfaces, memory, tool and skill registries, and routing or workflow
controllers \citep{chen2026harnessxcomposableadaptiveevolvable,gu2026modelscalingscalingscaling}. Prior systems instantiate these functions through reasoning--action loops, tool coordination, persistent memory, reflection, and executable skills
\citep{karpas2022mrklsystemsmodularneurosymbolic,
packer2024memgptllmsoperatingsystems,schick2023toolformerlanguagemodelsteach,
shen2023hugginggptsolvingaitasks,shinn2023reflexion,voyager,yao2023react}.
Harness engineering increasingly revises such contents from execution feedback, including prompts, memories, skills, workflows, and cross-component repairs
\citep{abuzakuk2026optimizingagenticworkflowsusing,
chen2026failedtrajectoriesreliablellm,
liu2026adaptiveautoharnesssustainedselfimprovement,
yao2026harnessbenchmeasuringharnesseffects,
zhang2026selfharnessharnessesimprove,
zhang2026skillflowbenchmarkinglifelongskilldiscovery,
zhong2026skilllearnbenchbenchmarkingcontinuallearning,
zhou2023largelanguagemodelshumanlevel}. 
However, the main objective of these methods is usually the quality of a component or the next configuration on a current task or target distribution. Repeated improvement alone does not provide a general retention criterion for the full harness state  \citep{lin2026harnessupdatingharnessbenefit}. HCL instead treats the mutable harness as a unified continual learning state and evaluates retention across committed updates.

\subsection{Model-Centric Continual Learning}

Model-centric continual learning adapts a model to a non-stationary stream of tasks or data while seeking to retain capabilities acquired from earlier experience.  Its central challenge is catastrophic forgetting, which arises when learning new knowledge disrupts knowledge acquired  by the model \citep{delange2021continual,kirkpatrick2017overcoming,
wang2024comprehensive}. Model-centric continual learning addresses catastrophic forgetting through replay, regularization, architectural isolation, representation learning, and constrained optimization\citep{abbes2026replaygradient,bellitto2024saliencyreplay,
kangdon,lewandowski2025spectralregularization,
liu2026cp,lopezpaz2017gem,lu2024revisitingarchitectures,
shang2025divideorthogonalize,urettini2025ocar,
wang2022dualpromptcomplementarypromptingrehearsalfree,
wang2022l2p,wang2025cutreplay,yue2025tdgr}. Recent work extends these families to large language models and broader knowledge streams, but the evolving state remains model knowledge, representations, architectures, or parameters \citep{zhao2026on,zhang2026multi}.
HCL transfers the same acquisition of new capability and retention of old capability problem to the state outside a frozen model. 

\section{Harness Continual Learning}
\label{sec:hcl_framework}

\subsection{Definition and Problem Setting}
\label{sec:problem_setting}

Consider a fixed foundation model $F_{\theta}$ and a harness $H_n$ deployed at
interaction step $n$. The model parameters $\theta$ remain unchanged.
We define \emph{Harness Continual Learning} as as the problem of sequentially updating the deployed harness to acquire new behavior while retaining behavior that was reliable before the update. Such behavior may be a correct response, a valid tool call, or an action
trajectory satisfying an environment goal. Retention requires such behavior to remain successful after
later harness updates when evaluated under the same input and execution
conditions.

At interaction step $n$, $\mathbf{u}_n$ denotes the raw interaction, such as an
instruction, an observation, or a multimodal input. The harness transforms
$\mathbf{u}_n$ into the structured interaction $\mathbf{i}_n$. Guided by the
frozen foundation model, it then combines $\mathbf{i}_n$ with selected memory
and capabilities to assemble the execution context $\mathbf{z}_n$. The model
and external runtime execute $\mathbf{z}_n$ to produce the outcome
$\mathbf{y}_n$. Post-execution feedback is denoted by $\mathbf{f}_n$. We
collect these interaction-level objects as
\begin{equation}
\label{eq:interaction_evidence}
\mathbf{e}_n=(\mathbf{u}_n,\mathbf{i}_n,\mathbf{z}_n,\mathbf{y}_n,\mathbf{f}_n).
\end{equation}
The Optimizer provides the foundation model $F_{\theta}$ with an update rule, the deployed harness, and the available interaction evidence as context for generating a candidate harness,
\begin{equation}
\label{eq:hcl_candidate_transition}
\widetilde{H}_{n+1}=\mathcal{O}_{F_{\theta}}(H_n,\mathbf{e}_n),
\end{equation}
The candidate remains separate from the deployed harness until a commitment decision is made. Let $G_n \in \{0, 1\}$ denote this decision. The deployed harness evolves as:
\begin{equation}
\label{eq:hcl_general_transition}
H_{n+1}=\begin{cases}
\widetilde{H}_{n+1}, & G_n=1,\\
H_n, & G_n=0.
\end{cases}
\end{equation}

\begin{figure*}[t]
\centering
\includegraphics[width=\linewidth]{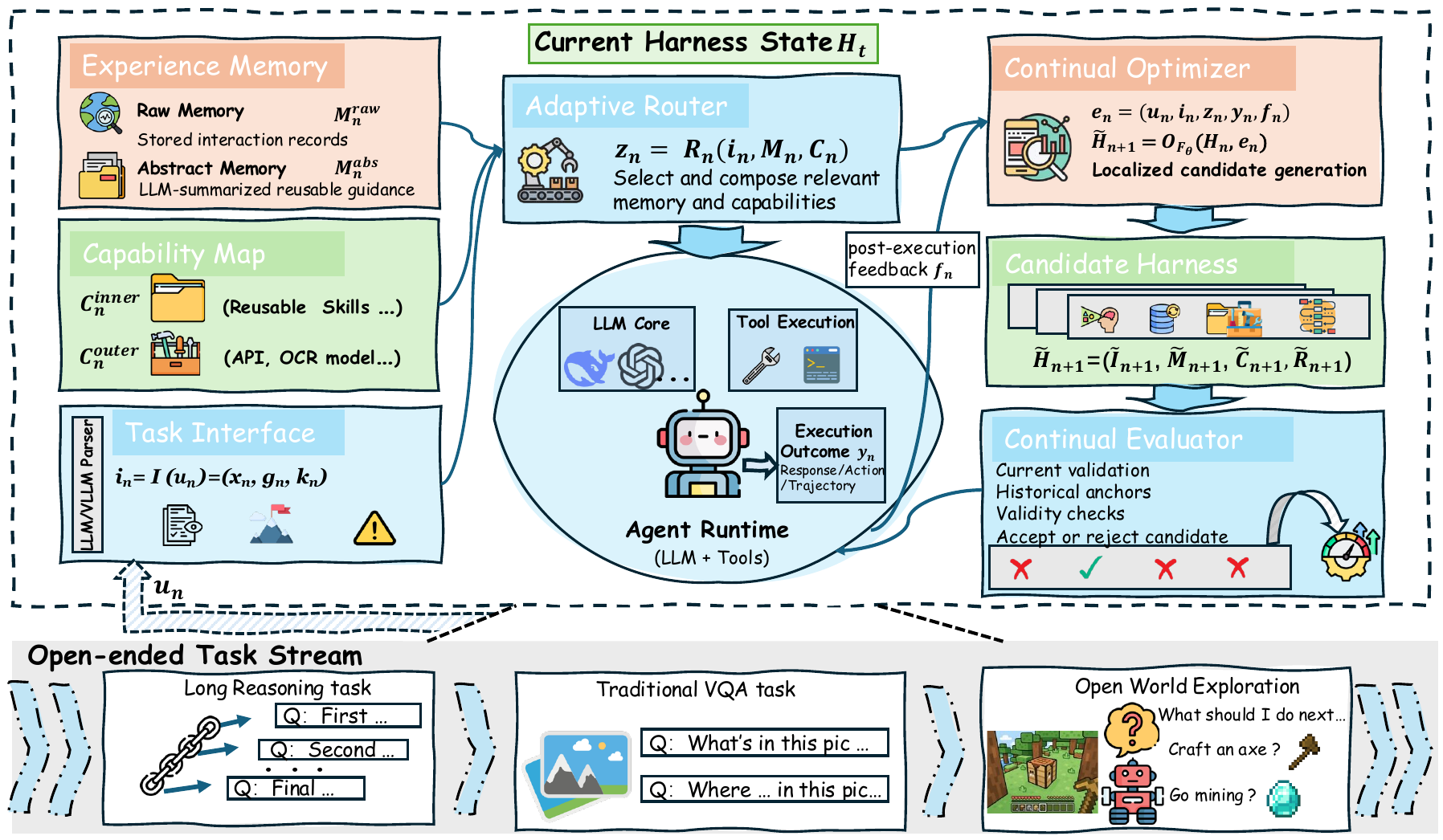}
\caption{Overview of HCL. The harness $H_n$ supports execution from interaction $\mathbf{u}_n$ to outcome $\mathbf{y}_n$. After execution, the Continual Optimizer proposes candidate harnesses, which the Continual Evaluator accepts or rejects based on current improvement, historical retention, and validity.}
\label{fig:hcl_framework}
\end{figure*}

\subsection{Harness State for Continual Learning}
\label{sec:evolving_state}

HCL organizes recurring execution functions into four jointly versioned components,
\begin{equation}
\label{eq:hcl_state}
H_n=(I_n,M_n,C_n,R_n),
\end{equation}
where $I_n$, $M_n$, $C_n$, and $R_n$ denote the Task Interface, Experience
Memory, Capability Map, and Adaptive Router. The four components are not meant
as a new decomposition of agent systems; they collect recurring execution functions into a state that can be versioned, evaluated, and updated under explicit acquisition
and retention constraints.
Figure \ref{fig:hcl_framework} shows how these components support execution and how the Continual Optimizer proposes candidate harnesses after execution, which are then evaluated by the Continual Evaluator before commitment.
\paragraph{Task Interface.}
The Task Interface transforms a raw interaction $\mathbf{u}_n$ into a structured
representation of the task input, objective, and execution constraints:
\begin{equation}
\mathbf{i}_n
=
I_n\left(\mathbf{u}_n\right)
=
\left(
\mathbf{x}_n,
\mathbf{g}_n,
\mathbf{k}_n
\right),
\end{equation}
where $\mathbf{x}_n$ denotes the available input,
$\mathbf{g}_n$ the task objective, and
$\mathbf{k}_n$ constraints such as output format, legal tool use, and
environment restrictions. Internally, $I_n$ includes the prompts, task
templates, and parsing or normalization rules used to construct this
representation. Since changes to these contents may alter how future tasks are
interpreted, $I_n$ is versioned as part of the evolving harness.

\paragraph{Experience Memory.}
Agent memory can take many forms, including episodic records, summaries, and
reflections
\citep{
packer2024memgptllmsoperatingsystems,
park2023generativeagentsinteractivesimulacra,
shinn2023reflexion,
wang2024agentworkflowmemory,
zhong2024memorybank}. HCL organizes accumulated experience as
$M_n=(M_n^{\mathrm{raw}},M_n^{\mathrm{abs}})$, where
$M_n^{\mathrm{raw}}$ and $M_n^{\mathrm{abs}}$ denote Raw Memory and Abstract
Memory, respectively.

Raw Memory $M_n^{\mathrm{raw}}$ stores the task input $\mathbf{u}_n$, the
resulting response or action trajectory $\mathbf{y}_n$, and the corresponding
environment or verifier feedback $\mathbf{f}_n$. To keep storage bounded, it
retains a fixed number of interactions from each task in arrival order,
preserving evidence of successful behavior and encountered failures.
Abstract Memory $M_n^{\mathrm{abs}}$ summarizes Raw Memory into reusable, such as output conventions, reliable reasoning patterns, and
common errors to avoid. As new interactions accumulate, abstract entries can be
added or updated to support transfer across related tasks.

Together, Raw Memory preserves concrete experience for reuse and recovery,
while Abstract Memory provides reusable knowledge for cross-task adaptation.
\paragraph{Capability Map.}
The Capability Map defines the operations and skills available to the agent
during execution. We denote it as
$C_n=(C_n^{\mathrm{outer}}, C_n^{\mathrm{inner}})$, where
$C_n^{\mathrm{outer}}$ contains externally provided capabilities and
$C_n^{\mathrm{inner}}$ contains skills acquired through continual interaction.

Outer capabilities connect the frozen model to external resources, such as
APIs, retrieval services, calculators, and environment
actions. Each entry specifies its function, invocation interface, applicable
conditions, and limitations.
Inner capabilities are reusable skills abstracted from
$M_n^{\mathrm{abs}}$. An LLM can consolidate related abstract memories into
procedures with explicit inputs, outputs, execution steps, and applicable
scopes. As Abstract Memory evolves, new inner skills can be added and existing
ones revised.
Unlike a static capability library, $C_n$ can expand through experience,
allowing the frozen-model agent to continually acquire and reuse executable
skills across tasks.

\paragraph{Adaptive Router.}
The Adaptive Router connects the Task Interface, Experience Memory, and Capability Map to task execution. Given the structured interaction $\mathbf{i}_n$, it retrieves relevant experience from $M_n$, selects capabilities from $C_n$, and organizes them into an execution context:
\begin{equation}
\label{eq:hcl_execution}
\mathbf{z}_n=R_n(\mathbf{i}_n,M_n,C_n).
\end{equation}
Its routing prompts, selection criteria, and workflow templates may evolve as
memory and capabilities change. 
Detailed storage and update boundaries are given in
Appendix~\ref{app:harness_details}.

\subsection{Guarded Harness Evolution}
\label{sec:guarded_evolution}

\paragraph{Continual Optimizer.}
Guarded harness evolution separates candidate generation from deployment. From $H_n$ and $\mathbf{e}_n$, the Continual Optimizer diagnoses which harness contents may explain the observed outcome and proposes revisions to the Task Interface, Experience Memory, Capability Map, or Adaptive Router state. When several components are selected, we use a sequential component-wise search: components are considered in a predefined order, up to $K$ alternatives are generated for the current component while the remaining state is fixed, and the highest-scoring admissible alternative becomes the basis for the next component. If no alternative passes the gate, that component is left unchanged. The deployed
$H_n$ is never modified during this search; only the final candidate can be committed. 

\paragraph{Continual Evaluator.}
The Continual Evaluator E examines three complementary aspects: \emph{current improvement} measures whether the candidate better solves the current task, \emph{historical retention} checks whether previously reliable behavior is preserved, and \emph{validity} ensures that the updated harness and its outputs remain usable. Let $V_n$ be current-task validation cases and $P(H,V_n)$ the corresponding performance. Current improvement is
\begin{equation}
\label{eq:evaluator_validation}
\Delta_n=P(\widetilde{H}_{n+1},V_n)-P(H_n,V_n).
\end{equation}
For historical retention, the Continual Evaluator maintains an evaluation-only anchor set
$A_n$. 
Anchors preserve raw inputs and task-specific success criteria from previously observed examples chosen by an LLM-based selector (Appendix~\ref{app:anchor-set}).
They are unavailable to execution and candidate generation, serving only
as a retention test.
For each anchor $a$, define
\begin{equation}
\label{eq:anchor_success}
q(H,a)\in\{0,1\},
\end{equation}
where success is task-specific. Historical loss counts anchors solved by the
current harness that fail under the candidate:
\begin{equation}
\label{eq:evaluator_historical_loss}
D_n=\sum_{a\in A_n}\mathbf{1}\!\left[
q(H_n,a)=1\land q(\widetilde{H}_{n+1},a)=0\right].
\end{equation}
The historical loss serves as an empirical proxy for forgetting by measuring
performance regression on sampled historical anchors.
Task-specific success criteria are listed in Appendix~\ref{app:anchor_criteria}. Let
$v_n(\widetilde H)\in\{0,1\}$ denote whether a candidate satisfies applicable
syntax, output-schema, tool-use, task, and environment constraints. Candidate
$k$ is admissible if
\begin{equation}
\label{eq:evaluator_decision}
G_n^{(k)}=\mathbf{1}\!\left[
\Delta_n^{(k)}\ge\delta_n\ \land\
D_n^{(k)}\le B_n\ \land\
v_n(\widetilde H_{n+1}^{(k)})=1
\right].
\end{equation}

This gate separates local proposal generation from persistent state change.
Under Eq.~(\ref{eq:evaluator_decision}), a candidate is admissible only if it
(i) improves current-task performance by at least $\delta_n$,
(ii) keeps historical regression on the sampled anchors within $B_n$, and
(iii) satisfies task-specific validity requirements.
If multiple candidates are admissible,
the Evaluator selects the candidate with the highest current-task performance,
using smaller anchor-based historical regression as a tie-breaker. If no
candidate is admissible, the currently deployed harness remains unchanged.


\subsection{Connections to Model-Centric Continual Learning}
\label{sec:hcl_model_cl}

HCL is related to several principles in model-centric continual
learning, but realizes them through harness evolution rather than parameter
updates \citep{delange2021continual,wang2024comprehensive}. Experience Memory
plays a role analogous to replay by retaining past interactions,
while the Capability Map corresponds to representation-based methods
by abstracting accumulated experience into reusable skills. The Adaptive
Router is analogous to architecture-based methods, as it dynamically selects
and composes capabilities into workflows.
Similarly, the Continual Optimizer and Continual Evaluator introduce an
update-and-retention mechanism analogous to optimization and regularization:
candidate harness changes are proposed from current feedback and committed
only when they improve current performance without violating historical
retention and validity constraints.

The key distinction is that HCL moves continual learning beyond the model
parameters themselves. Rather than treating replay, representation,
architecture, and update control as separate strategies for adapting a model
\citep{kang2025dynamic,liu2026branch}, HCL integrates their functions into a
single evolving harness under the same acquisition--retention objective.
This provides a system-level view of continual learning, where not only the
model but also the surrounding agent infrastructure becomes a persistent
carrier of continually acquired capabilities.

\section{Experiments}
\label{sec:experiments}

We evaluate HCL on ALFWorld \citep{alfworld} and Minecraft
\citep{voyager} to study long-horizon capability accumulation and reuse,
and on textual and multimodal task streams to measure harness-level
forgetting through repeated evaluation of earlier tasks.
We use frozen Qwen-family models for interactive and multimodal experiments
and frozen DeepSeek-family models for the main textual stream.
Further analyses examine stability--plasticity control, stronger-model
assistance for harness evolution, and individual component contributions.

Validation cases and historical anchors are used only for candidate
selection, while disjoint test sets are reserved for reporting.
The anchor set is updated after each task and remains fixed during
candidate generation and evaluation for the next task.
Stability-HCL ($B_n=0$) and Plasticity-HCL ($B_n=\infty$) differ only
in the historical-loss budget, with identical current-improvement
and validity requirements.
For stage-wise benchmarks, we report final average performance
(\textit{Final Avg.}) and average forgetting (\textit{Avg. Fgt.})
following standard continual learning protocols
\citep{delange2021continual,ref3}.
Forgetting measures the mean decline from each earlier task's best
observed performance to its final performance.
Full experimental settings, including anchor-set sizes and proposal
budgets, are provided in Appendix~\ref{app:experimental_settings}.

\subsection{Open-World Capability Accumulation}
\label{sec:open_world_experiments}

\paragraph{ALFWorld.}
We evaluate HCL on six sequential ALFWorld task categories using frozen
Qwen3.5-9B, with 10 training episodes per category and 134 evaluation
episodes. We compare against Static Harness, RAG, MemP \citep{memp},
and MemRL \citep{zhang2026memrl}, with the memory-based baselines
reimplemented under the same execution framework for fair comparison.
As shown in Table~\ref{tab:alfworld_main}, all methods improve over the Static Harness, while HCL achieves the strongest overall performance. Plasticity-HCL obtains the highest final average success of 62.98\%, compared with 55.56\% for the best non-HCL baseline, but incurs 10.94\% average forgetting. Stability-HCL achieves a comparable 61.74\% final average while reducing average forgetting to 2.64\%. Since the two HCL variants share the same frozen model and update mechanism and differ only in the historical-loss tolerance $B_n$, their contrast directly illustrates how the Continual Evaluator controls the stability--plasticity trade-off between capability acquisition and retention.
The particularly large gain on the Two-object category reflects HCL's ability to reuse and compose accumulated experience, skills, and routing patterns for longer-horizon multi-object interactions. In contrast, the lower performance on Cool relative to MemRL suggests that some task-specific behaviors may benefit more from episodic memory, while harness-level updates can introduce task-dependent interference.

\begin{table}[t]
\caption{Final task-wise performance on ALFWorld. The best and second-best results in each metric column are marked in \textbf{bold} and \underline{underlined}, respectively.}
\label{tab:alfworld_main}
\begin{center}
\footnotesize
\setlength{\tabcolsep}{3.1pt}
\resizebox{\linewidth}{!}{%
\begin{tabular}{lcccccc|cc}
\toprule
Method & Pick & Look & Clean & Heat & Cool & Two-object
& Final Avg. $\uparrow$ & Avg. Fgt. (\%) $\downarrow$ \\
\midrule
Static Harness
& {95.80}
& 66.70
& 25.80
& 26.10
& 9.50
& 58.80
& 47.12
& -- \\

RAG Baseline
& {95.80}
& \underline{83.30}
& 41.90
& \underline{39.10}
& 14.30
& 58.80
& 55.56
& \textbf{1.74} \\

MemP \citep{memp}
& {95.80}
& \underline{83.30}
& \underline{48.40}
& 34.80
& 9.50
& 47.10
& 53.15
& 5.18 \\

MemRL \citep{zhang2026memrl} & \underline{95.83} & \textbf{88.89} & 41.94 & \textbf{39.13} & \textbf{57.14} & 41.18 & 60.68 & \underline{1.76} \\

\textbf{Stability-HCL (Ours)}
& \textbf{100.00}
& \underline{83.30}
& \textbf{51.60}
& 30.40
& \underline{28.60}
& \underline{76.50}
& \underline{61.74}
& {2.64} \\

\textbf{Plasticity-HCL (Ours)}
& \textbf{100.00}
& {77.80}
& 41.90
& \underline{39.10}
& 19.00
& \textbf{100.00}
& \textbf{62.98}
& 10.94 \\
\bottomrule
\end{tabular}}
\end{center}
\end{table}

\paragraph{Minecraft.}
\begin{figure*}[t]
    \centering
    \includegraphics[width=\textwidth]{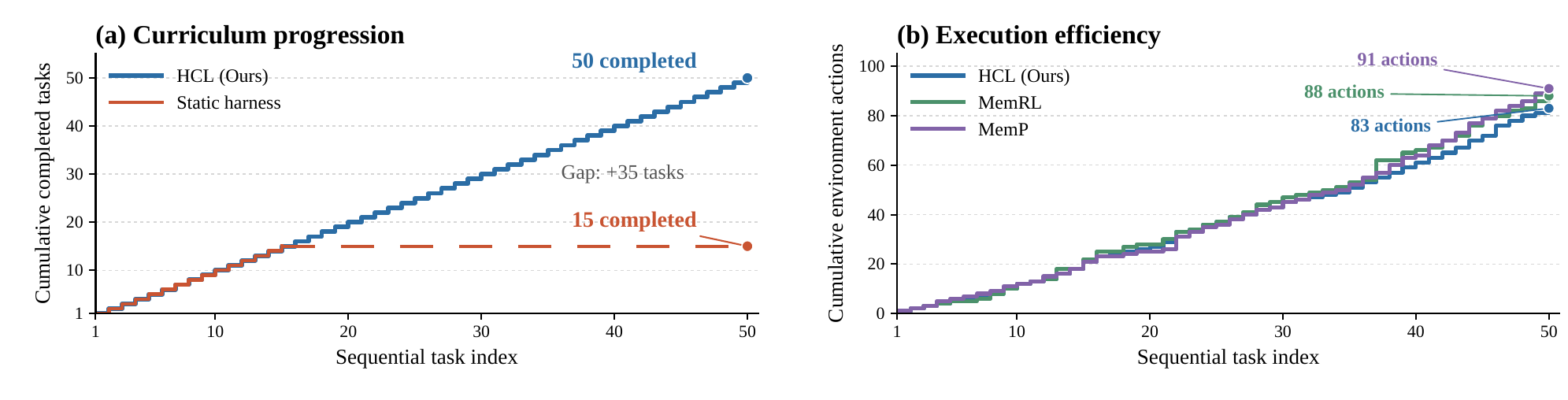}
    \caption{Minecraft curriculum progression and execution efficiency. 
    {(a)} Plasticity-HCL completes all 50 tasks, while the Static Harness plateaus at
    15. {(b)} Plasticity-HCL uses 83 cumulative environment actions, versus 88 for
    MemRL and 91 for MemP; lower is more efficient.}
    \label{fig:minecraft_evolution}
\end{figure*}

\begin{table}[t]
\caption{Final task-wise performance on the textual and multimodal streams.
}
\label{tab:controlled_main}
\begin{center}
\footnotesize
\setlength{\tabcolsep}{3.1pt}
\resizebox{\linewidth}{!}{%
\begin{tabular}{lcccc|cc}
\toprule
\multicolumn{7}{l}{\textit{Textual reasoning: MuSiQue $\rightarrow$ ProofWriter $\rightarrow$ GSM8K $\rightarrow$ HotpotQA}} \\
Method & MuSiQue & ProofWriter & GSM8K & HotpotQA & Final Avg. $\uparrow$ & Avg. Fgt. (\%) $\downarrow$ \\
\midrule
DeepSeek-V4.1-Flash Zero-shot & 48.00 & 79.20 & 65.20 & 62.20 & 63.65 & -- \\
MemP \citep{memp} & 51.40 & 78.80 & 70.80 & 60.80 & 65.45 & 0.47 \\
MemRL \citep{zhang2026memrl} & \underline{53.80} & 80.00 & 77.40 & \textbf{64.20} & 68.85 & \textbf{0.07} \\
\textbf{Plasticity-HCL (Ours)} & \textbf{58.60} & \textbf{85.60} & \textbf{96.00} & \textbf{64.20} & \textbf{76.10} & 0.40 \\
\textbf{Stability-HCL (Ours)} & 53.60 & \underline{82.80} & \underline{82.60} & \underline{63.60} & \underline{70.65} & \underline{0.27} \\
\midrule
\multicolumn{7}{l}{\textit{Multimodal perception: Detection $\rightarrow$ Caption $\rightarrow$ Grounding $\rightarrow$ VQAv2}} \\
Method & Detection & Caption & Grounding & VQAv2 & Final Mean $\uparrow$ & Avg. Fgt. (\%) $\downarrow$ \\
\midrule
Qwen3.6-27B Zero-shot & 37.52 & 21.10 & 43.00 & \textbf{85.47} & 46.77 & -- \\
MemP \citep{memp} & 58.67 & 31.89 & 90.80 & 79.27 & 65.16 & \textbf{0.00} \\
MemRL \citep{zhang2026memrl} & 59.33 & 30.66 & \underline{91.40} & 77.60 & 64.75 & \textbf{0.00} \\
\textbf{Plasticity-HCL (Ours)} & \underline{64.14} & \underline{37.31} & 90.60 & \underline{79.80} & \underline{67.96} & 0.81 \\
\textbf{Stability-HCL (Ours)} & \textbf{65.34} & \textbf{39.41} & \textbf{91.60} & 79.33 & \textbf{68.92} & \underline{0.22} \\
\bottomrule
\end{tabular}}
\end{center}
\end{table}
We further examine long-horizon evolution in Minecraft using deepseek-v4-flash-0731 on a 50-task curriculum spanning collection, crafting, mining, tool
use, placement, smelting, and multi-step dependencies. Environment feedback is
stored and can revise reusable capabilities and workflows, while retained skill
tests serve as historical anchors. We use Plasticity-HCL in this setting, while MemRL and MemP are reproduced within the same harness stack as baselines.
Figure~\ref{fig:minecraft_evolution} shows that the Static Harness follows HCL
for 15 tasks and then plateaus, whereas HCL completes all 50. HCL uses 83
environment actions, compared with 88 for MemRL and 91 for MemP, indicating
less redundant execution while retaining progression across the full
curriculum. This setting complements the stage-wise benchmarks by testing
capability accumulation and reuse over a substantially longer interaction
sequence.

\subsection{Controlled-Stream Continual Learning}
\label{sec:data_stream_experiments}
\label{sec:main_results}

\paragraph{Textual and multimodal streams.}
We evaluate HCL on two controlled sequential task streams.
The textual stream follows the sequence of MuSiQue
\citep{trivedi-etal-2022-musique}, ProofWriter
\citep{tafjord-etal-2021-proofwriter}, GSM8K
\citep{cobbe2021trainingverifierssolvemath}, and HotpotQA
\citep{yang-etal-2018-hotpotqa}. The multimodal stream consists of
COCO detection, COCO captioning, RefCOCO grounding, and VQAv2,
in that order.
We use frozen DeepSeek-V4.1-Flash for the textual stream and frozen
Qwen3.6-27B for the multimodal stream. Each task uses 250 adaptation
examples, 50 validation examples, and 500 test examples.
As shown in Table~\ref{tab:controlled_main}, HCL improves final average
performance on both streams, while Stability-HCL and Plasticity-HCL
exhibit different patterns of capability acquisition and retention.

On the textual stream, Plasticity-HCL achieves the highest final average of
76.10\% with 0.40\% average forgetting, while Stability-HCL ranks second with
a final average of 70.65\%. These results show that HCL achieves the strongest
overall performance on the textual stream. On the multimodal stream, HCL also
achieves the strongest overall balance between final performance and retention:
Stability-HCL achieves the best final mean of 68.92 with only 0.22\% average
forgetting, while Plasticity-HCL reaches 67.96 with 0.81\% forgetting. These
results show that HCL can accumulate capabilities across heterogeneous task
streams, while the historical-retention constraint controls the
stability--plasticity trade-off between adaptation and retention.

\subsection{Harness Evolution in Harness Continual Learning}
\label{sec:harness_evolution}

\paragraph{Stability--Plasticity Control}
\label{sec:stability_plasticity}

To examine the effect of historical-retention constraints, we vary the
historical-loss budget $B_n\equiv b$ over $b\in\{0,1,3,\infty\}$
using frozen DeepSeek-V4-Flash, while keeping the current-improvement
and validity criteria fixed. Each previous task contributes 80 historical
anchors. For each candidate update, $b$ limits the total number of anchors
that are solved by the deployed harness but fail under the candidate.
Thus, $b=0$ requires all currently solved anchors to remain successful,
whereas $b=\infty$ removes this constraint. Additional experimental
settings and stage-wise results are provided in
Appendix~\ref{app:stability_analysis}.
As shown in Table~\ref{tab:text_threshold_main}, $b=0$ yields the lowest
average forgetting of 0.39\%, whereas $b=1$ achieves the highest final
average accuracy of 63.46\%, improving over $b=0$ by 2.21 percentage
points with average forgetting of 1.22\%.
Increasing $b$ to $3$ and $\infty$ reduces final average accuracy to
62.04\% and 60.13\%, while average forgetting rises to 2.00\% and
3.45\%, respectively.
These results show that limited tolerance for historical regression
improves final average accuracy in this setting, but further relaxation
reduces both final accuracy and retention.
The budget constrains regression on sampled anchors rather than all
historical test cases, so $b=0$ does not guarantee zero test-set forgetting.

\begin{table}[t]
\caption{
Effect of historical-loss tolerance $b$ on final performance and forgetting.
}
\label{tab:text_threshold_main}
\begin{center}
\footnotesize
\setlength{\tabcolsep}{6pt}
\begin{tabular}{lcccc|cc}
\toprule
$b$ & MuSiQue & ProofWriter & GSM8K & HotpotQA
& Final Avg. $\uparrow$ & Avg. Fgt. (\%) $\downarrow$ \\
\midrule
$0$ & \underline{27.83} & 73.33 & \underline{84.33} & \textbf{59.50}
& 61.25 & \textbf{0.39} \\
$1$ & 24.83 & \underline{77.50} & \textbf{92.33} & \underline{59.17}
& \textbf{63.46} & \underline{1.22} \\
$3$ & 26.83 & \textbf{79.83} & 83.00 & 58.50
& \underline{62.04} & 2.00 \\
$\infty$ & \textbf{28.33} & 71.00 & 82.00 & \underline{59.17}
& 60.13 & 3.45 \\
\bottomrule
\end{tabular}
\end{center}
\end{table}

\paragraph{External-Model Assistance}
\label{sec:harness_distillation}

We further examine whether a stronger external model can improve HCL without replacing the frozen model. 
Base-HCL uses Qwen3-4B throughout. Opt-HCL uses Qwen3.6-27B only to
generate candidate harness updates, while Qwen3-4B handles online
execution. Online-HCL instead uses Qwen3.6-27B for per-example task
structuring, workflow planning, and memory, skill, and tool selection.
All variants follow Stability-HCL ($B_n=0$), with all other
settings held fixed.
\begin{figure*}[t]
    \centering
    \includegraphics[width=\linewidth]
    {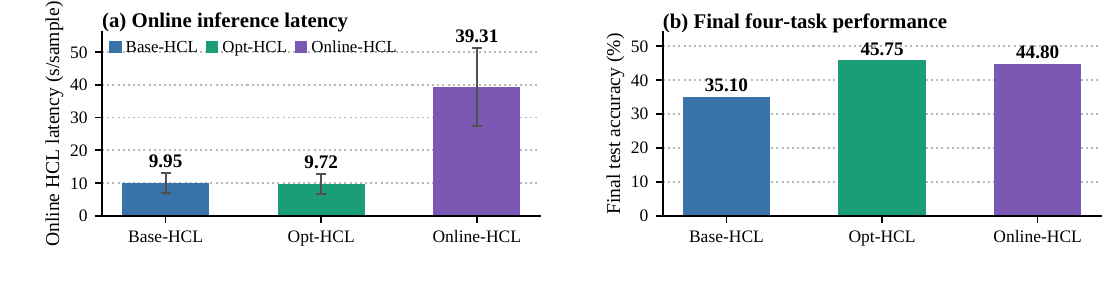}
    \caption{
        Effect of stronger-model placement in HCL.
        Base-HCL is fully driven by Qwen3-4B.
        Opt-HCL uses Qwen3.6-27B only for candidate harness generation while
        keeping Qwen3-4B for online execution.
        Online-HCL uses Qwen3.6-27B for online HCL operations.
        (a) Mean online processing time per matched sample with one standard
        deviation.
        (b) Final four-task average accuracy.
    }
    \label{fig:harness_distillation_result_v2}
\end{figure*}
As shown in Figure~\ref{fig:harness_distillation_result_v2}, Opt-HCL
achieves a final average accuracy of 45.75\%, improving over Base-HCL
(35.10\%) by 10.65 percentage points while maintaining comparable online
latency (9.72 vs.\ 9.95 seconds per sample).
Online-HCL also improves over Base-HCL, reaching 44.80\% accuracy,
but its online latency is substantially higher at 39.31 seconds per sample.
In Opt-HCL, Qwen3.6-27B is used only to generate candidate harness updates;
accepted updates are retained for subsequent online execution with the
frozen Qwen3-4B model.
These results suggest that, in this setting, stronger-model assistance
can improve HCL through persistent harness updates without requiring
the stronger model during online execution.

\subsection{Ablation Study}
\label{sec:component_ablation}

We conduct component ablations to examine the contribution of each HCL
component. Specifically, on the controlled multimodal stream with frozen
Qwen3.5-4B, we keep one component fixed at a time while allowing the other
three to continue adapting. The fixed component remains available with its
initialized contents, and all variants share the same data allocation,
evaluation criteria, and update schedule. Interventions and full per-task
results are provided in Appendix~\ref{app:component_ablation_details}.
As shown in Table~\ref{tab:component_ablation}, Plasticity-HCL achieves the highest
final average of 63.41\%, compared with 62.28--63.12\% for restricted
variants. Fixing Memory causes the largest performance drop and increases
average forgetting to 0.83\%. Some restricted variants exhibit lower forgetting
because they also acquire less new capability. These results indicate that all
four components contribute to continual adaptation, and that forgetting should
be interpreted together with capability acquisition rather than in isolation.

\begin{table}[t]
\caption{Component ablation on the controlled multimodal stream. A component is
kept available during execution but its persistent updates are frozen.}
\label{tab:component_ablation}
\begin{center}
\footnotesize
        \begin{tabular}{lcccc|cc}
            \toprule
            Component
            & $I$
            & $M$
            & $C$
            & $R$
            & Final Avg. $\uparrow$
            & Avg. Fgt. (\%) $\downarrow$ \\
            \midrule

            Zero-shot
            & -- & -- & -- & --
            & 34.84
            & -- \\

            w/o Interface update
            & $\times$ & $\checkmark$ & $\checkmark$ & $\checkmark$
            & 62.37
            & \underline{0.11} \\

            w/o Memory update
            & $\checkmark$ & $\times$ & $\checkmark$ & $\checkmark$
            & 62.28
            & 0.83 \\

            w/o Capability update
            & $\checkmark$ & $\checkmark$ & $\times$ & $\checkmark$
            & \underline{63.12}
            & \textbf{0.06} \\

            w/o Router update
            & $\checkmark$ & $\checkmark$ & $\checkmark$ & $\times$
            & 62.77
            & 0.14 \\

            Plasticity-HCL
            & $\checkmark$ & $\checkmark$ & $\checkmark$ & $\checkmark$
            & \textbf{63.41}
            & 0.45 \\

            \bottomrule
        \end{tabular}
\end{center}
\end{table}

\section{Conclusion}
\label{sec:conclusion}

We formulate Harness Continual Learning (HCL) as a new continual learning paradigm
in which the agent harness, rather than model parameters, evolves through
sequential experience. HCL represents mutable harness components as a unified
continual state and introduces a proposal--evaluation--commitment mechanism to
balance capability acquisition and historical retention.
Experiments across open-world interaction, textual reasoning, and multimodal
perception show that frozen models can continually acquire and retain
capabilities through the proposed harness evolution. Further analysis reveals the stability--plasticity trade-off of harness
adaptation and shows that stronger models provide a more efficient strategy
when used for candidate generation rather than continuous online execution. These results extend
continual learning from parameter adaptation to the evolution of
agent-level infrastructure, establishing harness evolution as a complementary
dimension of continual learning.

\subsection*{Reproducibility Statement}
Detailed model, data, evaluator, and reporting settings for the stage-wise
experiments are provided in Appendix~\ref{app:experimental_settings}.
Harness state boundaries and candidate-generation procedures are specified
in Appendix~\ref{app:harness_details}, while task-specific anchor criteria
are given in Appendix~\ref{app:anchor_criteria}. Additional settings for
the stability analysis, external capability extension, and component
ablations are provided in the corresponding appendices.

\subsection*{AI Use Statement}
Generative AI tools are used to assist with language editing. All technical content, experimental results, analyses, and claims are
reviewed and verified by the authors, who take responsibility for the final
content of the paper.

\bibliography{iclr2027_conference}
\bibliographystyle{iclr2027_conference}

\appendix
\section{Appendix}
The supplementary material provides detailed harness definitions and update
boundaries, detailed experimental settings and task-wise results, additional
stability--plasticity analyses, external capability details,
component ablations, and task-specific anchor criteria.

\section{Harness Details and Update Boundaries}
\label{app:harness_details}

\subsection{Harness and Evaluator Boundaries}
\label{app:implementation}

\paragraph{Task Interface.}
The Task Interface $I_n$ constructs $\mathbf{i}_n$ during execution. During
candidate generation, the Optimizer may revise its prompts, templates, and
parsing or normalization rules. Any such revision enters the deployed harness
$H_n$ only when the corresponding candidate is committed.

\paragraph{Raw and Abstract Memory.}
Raw and Abstract Memory $M_n^{\mathrm{raw}}$ and $M_n^{\mathrm{abs}}$ supply
interaction records and reusable guidance to the Router. The Optimizer may add
raw records or revise abstract entries during candidate generation, but these
changes become part of $H_n$ only after candidate commitment.

\paragraph{Capability Map.}
The Capability Map $C_n$ supplies the capabilities available to the Router.
The Optimizer may add or revise internal skills, while any resulting change is
excluded from the deployed state until the candidate containing it is
committed.

\paragraph{Adaptive Router.}
The Adaptive Router $R_n$ constructs $\mathbf{z}_n$. Its routing prompts,
selection criteria, and workflow templates may be revised by the Optimizer,
with accepted revisions entering $H_n$ only through candidate commitment.

\paragraph{Anchor Set.}
\label{app:anchor-set}
The Anchor Set $A_n$ is available only to the Evaluator and is unavailable to
both execution and candidate generation. After each adaptation batch, an LLM-based selector chooses a configured number of anchors from the observed examples using their predictions, correctness, format compliance, and task metadata. The selector prioritizes failures, format violations, brittle cases, and representative examples. Invalid or incomplete selections are completed by a deterministic fallback, after which the anchors are deduplicated and retained under a fixed per-task capacity. Historical anchors remain fixed during the optimization of subsequent tasks.

Thus, $H_n$ contains only persistent execution-time contents;
$\mathbf{i}_n$, $\mathbf{z}_n$, and $\mathbf{y}_n$ are transient, and $A_n$
remains evaluation-only.

\subsection{Optimizer Candidate Generation}
\label{app:optimizer_details}
Interaction feedback indicates whether an execution is successful but not how
the harness should change. In the implementation, memory and skill writes are
first evaluated jointly as a single implicit update and are either retained or
rolled back. The explicit candidate-generation stage then optimizes a
configuration-defined sequence of prompt and context artifacts, including the
task-interface structuring prompt and the router's workflow, memory-selection,
and context prompts. The task-interface artifact is skipped when no current
validation example invokes model-based structuring.

For each applicable artifact, the configured proposal model receives the
current artifact, post-execution metrics, current validation statistics,
aggregate historical-anchor statistics, and the commit-gate configuration. A
single generation call returns up to $K$ alternatives. The returned artifacts
are parsed and normalized, after which each alternative is evaluated
separately by temporarily replacing only the target artifact. Candidates that
fail artifact validation, do not change any comparable behavior, fail to meet
the required current-task improvement, exceed the historical-loss budget, or
violate the format-compliance threshold are rejected. Among the admissible
candidates, the Evaluator prioritizes current-task performance and uses lower
anchor loss as the first tie-breaker. If no candidate is admissible, the
artifact remains unchanged. Otherwise, the selected artifact is persisted and
applied immediately, so each subsequent artifact is generated and evaluated
against the already updated harness. Algorithm~\ref{alg:optimizer} summarizes
this procedure.

\begin{algorithm}[H]
\caption{Sequential Prompt-Artifact Optimization}
\label{alg:optimizer}
\begin{algorithmic}[1]
\Require deployed harness $H$, post-execution result $R_{\mathrm{train}}$,
configured proposal model $G$, current validation set $V$, historical anchors
$A$, configured artifact sequence $\mathcal{C}$, candidate budget $K$
\Ensure updated deployed harness $H$

\State $\mathcal{C}' \gets \textsc{ApplicableArtifacts}(\mathcal{C},V)$

\For{$c \in \mathcal{C}'$ \textbf{in configured order}}
    \State $R_0 \gets \textsc{Evaluate}(H,V,A)$
    \State $X_c \gets
    \textsc{GenerationContext}(H[c],R_{\mathrm{train}},R_0)$
    \State $\mathcal{Q}_c \gets
    \textsc{Normalize}\bigl(G(c,X_c,K)\bigr)$
    \Comment{one call returns at most $K$ alternatives}

    \For{$q \in \mathcal{Q}_c$}
        \State $H_q \gets \textsc{Replace}(H,c,q)$
        \State $R_q \gets \textsc{Evaluate}(H_q,V,A)$
    \EndFor

    \State $\mathcal{Q}_{\mathrm{ok}} \gets
    \{q \in \mathcal{Q}_c : \textsc{Gate}(q,R_q,R_0)=1\}$

    \If{$\mathcal{Q}_{\mathrm{ok}} \neq \varnothing$}
        \State $q^\star \gets
        \textsc{SelectBest}(\mathcal{Q}_{\mathrm{ok}},\{R_q\})$
        \State $H[c] \gets q^\star$
        \State \textsc{Persist}$(q^\star)$
        \Comment{commit before processing the next artifact}
    \EndIf
\EndFor

\State \Return $H$
\end{algorithmic}
\end{algorithm}

\section{Experimental Settings and Reporting Notes}
\label{app:experimental_settings}

\subsection{Experimental Settings}

The following paragraphs summarize the model, stream, adaptation and evaluator
data, and final reporting protocol for the stage-wise experiments retained in
this appendix. Unless otherwise stated, counts are per task or category.

Across these experiments, validation cases and historical anchors are restricted
to the Evaluator, and final test cases are used only for reporting. The main
Stability-HCL and Plasticity-HCL profiles use $B_n=0$ and $B_n=\infty$,
respectively. A main-profile candidate must improve by at least one validation
case for discrete metrics or strictly improve the designated continuous score,
without introducing an invalid outcome. 
For ALFWorld and Minecraft, because replaying candidate updates can alter the state of the interactive environment, we replace strict historical backtesting with an LLM-based evaluation that judges candidate acceptance given historical task records.
The independent textual sweep uses 40 proposal opportunities,
requires two additional correct predictions among 80 validation cases and at
least 90\% format compliance, and varies only $B_n\equiv b$ for
$b\in\{0,1,3,\infty\}$.

\paragraph{ALFWorld main.}
The stream contains six categories---Pick-and-Place, Look-in-Light, Clean,
Heat, Cool, and Two-object---and uses Qwen3.5-9B as the frozen model. Each
category provides 10 training episodes with at most 50 interaction steps per
episode, and the observed categories are evaluated after every stage. Final
reporting uses the 134 official evaluation episodes and includes category
macro-average success and average forgetting over the first five categories.

\paragraph{Minecraft main.}
This stream contains 50 tasks spanning collection, crafting, mining, tool use,
placement, smelting, and multi-step dependencies. HCL adapts from sequential environment feedback and uses retained
skill tests as historical anchors. We report cumulative task completion and
environment actions because completed tasks are not systematically replayed
after every update.

\paragraph{Textual main.}
The task order is MuSiQue $\rightarrow$ ProofWriter $\rightarrow$ GSM8K
$\rightarrow$ HotpotQA, with DeepSeek-V4.1-Flash frozen throughout. Thinking is
disabled and temperature is set to 0. Each task uses 250 adaptation examples,
50 validation examples, and 500 disjoint test examples. All methods share the
same model, task order, data allocation, and evaluation protocol. We report
normalized exact-match accuracy, its final task average, and average
forgetting.

\paragraph{Multimodal main.}
The task order is COCO detection $\rightarrow$ COCO captioning $\rightarrow$
RefCOCO grounding $\rightarrow$ VQAv2, with Qwen3.6-27B frozen throughout.
Each task uses 250 adaptation examples, 50 validation examples, and 500
disjoint test examples. Final reporting includes the task-specific scores,
their unweighted average, and forgetting.

\paragraph{Qwen textual comparison (appendix).}
The textual experiment follows the same MuSiQue, ProofWriter, GSM8K,
and HotpotQA order with Qwen3.6-27B frozen. Each task uses 250 adaptation
examples, 50 validation examples, and 500 disjoint test examples. It is
reported in Appendix~\ref{sec:textual_reasoning_experiment} as a cross-backbone
comparison and is not mixed with the main textual results.

\paragraph{Textual budget sweep.}
The independent sweep follows the same textual task order with
DeepSeek-V4-Flash. Each task uses 300 adaptation examples, 80 validation
examples, 80 anchors for every earlier task, and 600 test examples. The sweep
allows 40 proposal opportunities, ten at each task stage.

\paragraph{External-model assistance.}
The Base-HCL, Opt-HCL, and Online-HCL variants use the same task
stream, data allocation, update schedule, optimizer components, and
candidate budget. All three variants use the Stability-HCL acceptance
policy with $B_n=0$. Each candidate must yield at least one additional
correct validation prediction and must not forget any evaluated
historical anchor. The only experimental variable is the set of HCL
operations assigned to Qwen3.6-27B.

\paragraph{Capability extension.}
This experiment uses the same four reasoning tasks and a frozen Qwen3.6-27B
model, with an external arithmetic calculator registered in the Capability
Map. The evolving selector and Router determine whether the calculator is
invoked. We report final per-task accuracy and, for GSM8K, the number of
examples routed to the calculator.

\paragraph{Component ablation.}
The ablation follows the multimodal task order with Qwen3.5-4B frozen. Every
task uses 250 adaptation examples, 50 validation examples, and 500 test
examples; each variant keeps one persistent harness component fixed. We report
final per-task scores, average performance, forgetting, and committed-update
counts.

\subsection{Evaluation Metrics}
\label{sec:evaluation_metrics}

For the stage-wise ALFWorld, textual, and multimodal benchmarks, after stage
$s$ we evaluate the deployed harness on the current and all previously observed
tasks, obtaining $R_{s,j}$ for $j\leq s$. For streams whose task scores are
reported on a common 0--100 scale, we report
\begin{equation}
\label{eq:exp_metrics}
\operatorname{Avg}_T=\frac{1}{T}\sum_{j=1}^{T}R_{T,j},\qquad
\operatorname{Fgt}_T=\frac{1}{T-1}\sum_{j=1}^{T-1}
\left(\max_{r\in\{j,\ldots,T\}}R_{r,j}-R_{T,j}\right).
\end{equation}
For the multimodal stream, each task-specific reporting metric is expressed on
the 0--100 scale before taking the unweighted final mean and computing
forgetting, so forgetting is reported in percent (\%). Minecraft is
reported separately using cumulative task completion and environment actions,
because completed tasks are not systematically replayed after every update.
Stability-HCL sets $B_n=0$ and Plasticity-HCL $B_n=\infty$, with matched
current-improvement and validity criteria; $B_n$ constrains anchor loss rather
than test-set forgetting.

\subsection{Controlled Harness Continual Learning}
\label{sec:controlled_tasks}

We next evaluate HCL on task sequences. Within each stream, all HCL profiles
share the same foundation model, task order, data allocation, editable
artifacts, and candidate generator.

\subsubsection{Qwen Textual-Reasoning Experiment}
\label{sec:textual_reasoning_experiment}

For completeness, we retain the earlier Qwen3.6-27B textual experiment in this
appendix. The stream follows the order MuSiQue
\citep{trivedi-etal-2022-musique}, ProofWriter
\citep{tafjord-etal-2021-proofwriter}, GSM8K
\citep{cobbe2021trainingverifierssolvemath}, and HotpotQA
\citep{yang-etal-2018-hotpotqa}. These tasks cover multi-hop question
answering, logical deduction, mathematical reasoning, and
knowledge-intensive question answering. Qwen3.6-27B remains frozen throughout;
each task uses 250 adaptation, 50 validation, and 500 disjoint test examples.
The result is presented as a separate cross-backbone comparison and is not
pooled with the main textual experiment.

\begin{table}[t]
\caption{Textual-reasoning results with frozen Qwen3.6-27B. The best and second-best entries in each column are marked in \textbf{bold} and \underline{underlined}, respectively.}
\label{tab:text_qwen_appendix}
\begin{center}
\footnotesize
\setlength{\tabcolsep}{3.1pt}
\resizebox{\linewidth}{!}{%
\begin{tabular}{lcccc|cc}
\toprule
Method & MuSiQue & ProofWriter & GSM8K & HotpotQA
& Final Avg. $\uparrow$ & Avg. Fgt. (\%) $\downarrow$ \\
\midrule
Qwen3.6-27B Zero-shot & \textbf{57.00} & \textbf{83.80} & 44.80 & 62.40 & 62.00 & -- \\
MemP \citep{memp} & \underline{56.40} & 74.00 & \underline{54.00} & \textbf{65.80} & \underline{62.55} & \underline{0.20} \\
MemRL \citep{zhang2026memrl} & 55.80 & 75.20 & 52.20 & \underline{65.60} & 62.20 & \textbf{0.00} \\
\textbf{Plasticity-HCL (Ours)} & 56.00 & \underline{79.80} & \textbf{66.00} & 62.80 & \textbf{66.15} & 7.87 \\
\textbf{Stability-HCL (Ours)} & 55.80 & 77.00 & 51.00 & 64.20 & 62.00 & 1.27 \\
\bottomrule
\end{tabular}}
\end{center}
\end{table}

On Qwen3.6-27B, Plasticity-HCL attains the best final average (66.15\%) but
also substantially higher forgetting (7.87\%), whereas Stability-HCL trades
away adaptation for lower forgetting (1.27\%). This separate result indicates
that the realized stability--plasticity outcome depends on the frozen backbone
and optimization trajectory, even when the data budget and task order are
matched.

\subsection{Main-Result Reporting Notes}
The complete final task-wise results are reported directly in
Tables~\ref{tab:alfworld_main} and~\ref{tab:controlled_main}. Validation cases
and historical anchors are never used for final reporting; all reported main
scores use the disjoint test sets described above.

\section{Additional Stability--Plasticity Analysis}
\label{app:stability_analysis}

For this independent sweep, $\delta_n$ requires at least two additional correct
validation cases, every candidate must achieve at least 90\% output-format
compliance and introduce no syntax, tool-use, or environment violations, and
only $B_n\equiv b$ is varied. Each task uses 300 adaptation, 80 validation, and
600 test examples, with 80 anchors retained for every earlier task.

\begin{figure*}[t]
    \centering
    \includegraphics[width=\linewidth]
    {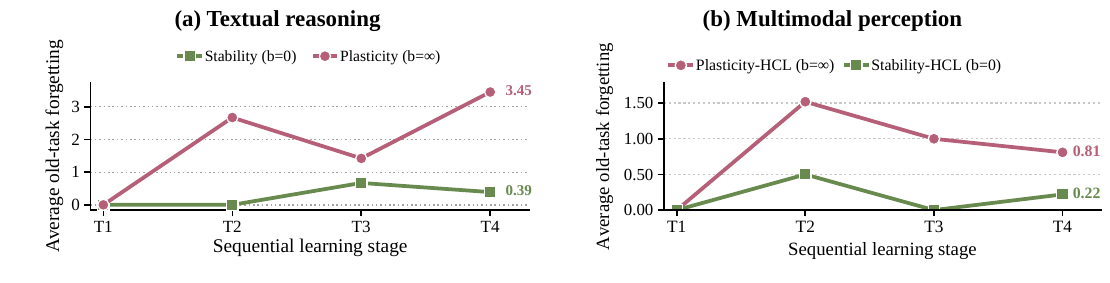}
    \caption{
        Stage-wise forgetting under different historical-loss tolerances.
        (a) Textual reasoning comparing Stability-HCL and Plasticity-HCL.
        Forgetting values are reported in percent.
        (b) Multimodal perception comparing Stability-HCL and Plasticity-HCL.
        Forgetting values are reported in percent.
    }
    \label{fig:controlled_forgetting}
\end{figure*}

The task-wise sweep results and stage-wise forgetting trajectories are shown in
Table~\ref{tab:text_threshold_main} and Figure~\ref{fig:controlled_forgetting}
in the main paper. The sweep is kept separate from the DeepSeek-V4.1-Flash
textual main experiment because it uses a different configuration and data
budget.

Across this sweep, final forgetting increases from
0.39\% at $b=0$ to 3.45\% at $b=\infty$. The multimodal trajectory in Figure~\ref{fig:controlled_forgetting} similarly shows 0.22\% rather than 0.81\% final forgetting for Stability-HCL versus Plasticity-HCL. Nonzero test forgetting at
$b=0$ reflects the finite coverage of the anchor set: preserving all currently
solved anchors does not guarantee unchanged behavior on historical test cases
not represented by $A_n$.

\section{External Capability Extension}
\label{app:capability_details}

We register an external arithmetic calculator in the Capability Map of a frozen
Qwen3.6-27B agent and allow the evolving selector and router to decide whether
it should be used for each input. Relative to an earlier Stability-HCL run, the
capability-aware run changes final accuracy from 58.60\% to 58.60\% on MuSiQue,
73.80\% to 82.80\% on ProofWriter, 53.20\% to 90.40\% on GSM8K, and 64.40\% to
65.80\% on HotpotQA. The largest change occurs on GSM8K, where 254 of 500 test
examples are routed to the calculator. Because the compared runs also differ in
prompts, retries, and memory trajectories, these changes do not isolate the
causal contribution of tool execution alone.

\begin{figure}[h]
    \centering
    \includegraphics[width=\linewidth]{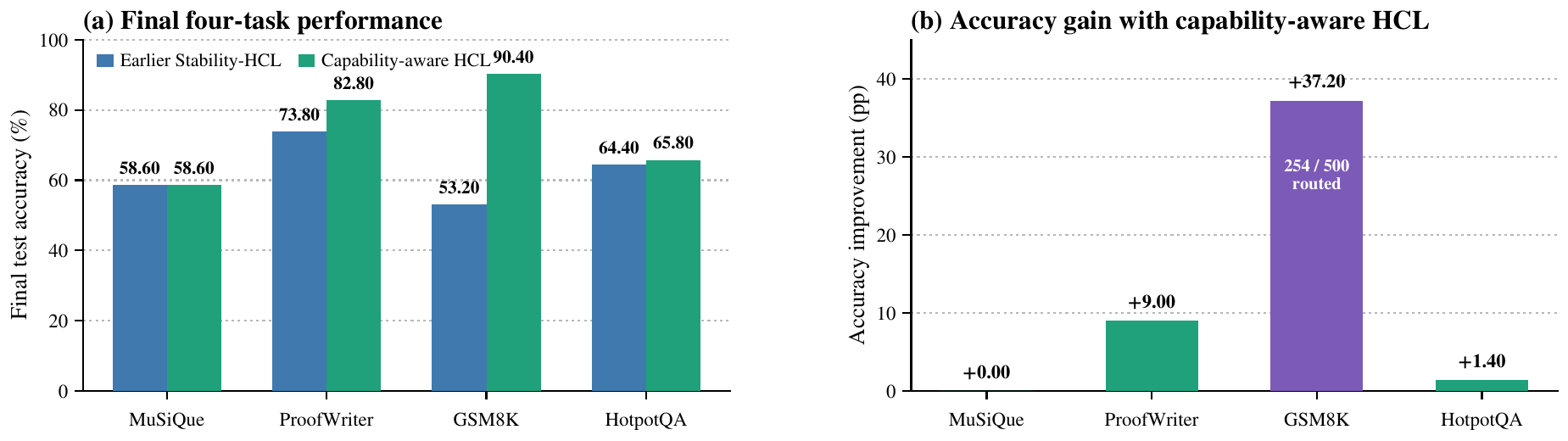}
    \caption{
    \textbf{Capability-aware HCL with an external arithmetic calculator.}
    (a) Final accuracy compared with the earlier Stability-HCL run.
    (b) Corresponding accuracy gains; GSM8K shows the largest improvement
    ($+37.20$ pp), with 254/500 examples routed to the calculator.
    The comparison is exploratory because the runs also differ in prompts,
    retries, and memory trajectories.
    }
    \label{fig:capability_aware_hcl}
\end{figure}

\section{Component Ablation Details}
\label{app:component_ablation_details}

All ablation variants use frozen Qwen3.5-4B and share the task order, data
allocation, evaluation criteria, and update schedule. A disabled component remains available during execution but retains its initialized contents throughout the stream. Zero-shot evaluates the frozen model without the structured HCL harness or
sequential updates.

\subsection{Ablation Configurations}

Table~\ref{tab:component_ablation_config} summarizes which harness components
are allowed to receive persistent updates in each variant.

\begin{table}[t]
\caption{
Update scope of the component-ablation variants. A $\checkmark$ permits
persistent updates, while $\times$ keeps the corresponding component fixed.
}
\label{tab:component_ablation_config}
\begin{center}
\small
\setlength{\tabcolsep}{3.5pt}
\begin{tabular}{L{0.24\textwidth}ccccL{0.44\textwidth}}
\toprule
Method & $I$ & $M$ & $C$ & $R$ & Fixed contents \\
\midrule
Zero-shot
& -- & -- & -- & --
& No structured HCL harness or persistent updates. \\

Full HCL
& $\checkmark$ & $\checkmark$ & $\checkmark$ & $\checkmark$
& None. \\

w/o Interface update
& $\times$ & $\checkmark$ & $\checkmark$ & $\checkmark$
& Prompts, templates, parsing, and normalization rules. \\

w/o Memory update
& $\checkmark$ & $\times$ & $\checkmark$ & $\checkmark$
& Raw and Abstract Memory entries. \\

w/o Capability update
& $\checkmark$ & $\checkmark$ & $\times$ & $\checkmark$
& Reusable skills. \\

w/o Router update
& $\checkmark$ & $\checkmark$ & $\checkmark$ & $\times$
& Routing prompts, selection criteria, and workflow templates. \\
\bottomrule
\end{tabular}
\end{center}
\end{table}

Because reusable skills may be derived from Abstract Memory, disabling Memory
updates also removes this source of newly acquired skills. The corresponding
variant therefore measures both direct memory adaptation and its downstream
effects on capability construction.

\subsection{Full Ablation Results}

Table~\ref{tab:component_ablation_full} reports the complete task-wise results
for the controlled multimodal stream, together with final average performance,
average forgetting.

\begin{table}[t]
\caption{
Full component-ablation results on the controlled multimodal stream.
``Committed'' counts candidate updates entering the persistent harness.
}
\label{tab:component_ablation_full}
\begin{center}
\small
\resizebox{\textwidth}{!}{
\begin{tabular}{lrrrrrrr}
\toprule
Method
& Detection
& Caption
& Grounding
& VQAv2
& Final Avg. $\uparrow$
& Avg. Fgt. (\%) $\downarrow$\\
\midrule

Zero-shot
& 35.11
& 22.98
& 0.00
& 81.27
& 34.84
& --
\\

w/o Interface update
& 53.45
& 33.56
& 87.80
& 74.67
& 62.37
& \underline{0.11}
\\

w/o Memory update
& 55.50
& 28.95
& 88.00
& 76.67
& 62.28
& 0.83
\\

w/o Capability update
& 55.11
& 34.16
& 86.40
& 76.80
& \underline{63.12}
& \textbf{0.06}
\\

w/o Router update
& 53.59
& 36.68
& 87.40
& 73.40
& 62.77
& 0.14
\\

Full HCL
& 53.07
& 36.09
& 87.60
& 76.87
& \textbf{63.41}
& 0.45
\\

\bottomrule
\end{tabular}
}
\end{center}
\end{table}

Full HCL achieves the highest final average score of 63.41, compared with
34.84 for the frozen zero-shot model. Restricting any individual component
reduces the final average, although the magnitude differs across components.
Freezing Interface updates reduces the final average to 62.37, with the main
drops appearing on Caption and VQAv2. Freezing Memory updates produces the
lowest ablated average, 62.28, and the largest average forgetting, 0.83\%,
indicating that adaptive memory is particularly important for retaining and
reusing experience across the stream. By comparison, disabling Capability
updates has a smaller effect on final performance (63.12), consistent with this
multimodal stream relying less on long-horizon executable skills than the
interactive Minecraft setting. Freezing Router updates yields 62.77 and notably
reduces VQAv2 performance to 73.40, suggesting that adaptive execution and
selection policies remain useful even when the other harness components can
continue to evolve.


\section{Anchor Success Criteria}
\label{app:anchor_criteria}

The fixed task-specific success criterion $q(H,a)$ in
Eq.~(\ref{eq:anchor_success}) is applied to the same raw input under $H_n$ and
$\widetilde{H}_{n+1}$.

\paragraph{Textual reasoning.}
For MuSiQue and HotpotQA, an anchor is successful when the normalized short
answer exactly matches an accepted reference answer. For ProofWriter, the
parsed entailment label must match the gold label and the output schema must be
valid. For GSM8K, the parsed final number must match the gold value after comma
and unit normalization.

\paragraph{Multimodal perception.}
For COCO detection, the queried category must be correct, the matched box must
have IoU $\ge 0.5$, and the box schema must be valid. For COCO captioning, the
sentence-level CIDEr score must be at least 0.5 on the normalized $[0,1]$
scale, and the caption schema must be valid. For RefCOCO grounding, the
predicted box must be valid and have IoU $\ge 0.5$ with the referred-object
box. For VQAv2, the standard VQA consensus score must be 1.0 after answer
normalization.

\paragraph{Interactive environments.}
For ALFWorld, the goal predicate must be reached within 50 steps under a valid
action sequence. For Minecraft, the retained skill test must reach its
predefined inventory or world-state predicate through a valid action sequence.

\end{document}

%% file: math_commands.tex
\usepackage{amsmath,amsfonts,bm}

\def\eqref#1{equation~\ref{#1}}

\def\1{\bm{1}}

\DeclareMathAlphabet{\mathsfit}{\encodingdefault}{\sfdefault}{m}{sl}
\SetMathAlphabet{\mathsfit}{bold}{\encodingdefault}{\sfdefault}{bx}{n}

